# Victim or Perpetrator?
# Analysis of Violent Character Portrayals from Movie Scripts

Victor R. Martinez[1], Krishna Somandepalli[1], Karan Singla[1],
Anil Ramakrishna[1], Yalda T. Uhls[2], Shrikanth Narayanan[1]

[1] University of Southern California, Los Angeles, CA. USA
[2] University of California Los Angeles, Los Angeles, CA, USA

**ABSTRACT**

Violent content in the media can influence viewers' perception of the society. For example, frequent depictions of certain demographics as victims or perpetrators of violence can shape stereotyped attitudes. We propose that computational methods can aid in the large-scale analysis of violence in movies. The method we develop characterizes aspects of violent content solely from the language used in the scripts. Thus, our method is applicable to a movie in the earlier stages of content creation even before it is produced. This is complementary to previous works which rely on audio or video post production. In this work, we identify stereotypes in character roles (i.e., victim, perpetrator and narrator) based on the demographics of the actor casted for that role. Our results highlight two significant differences in the frequency of portrayals as well as the demographics of the interaction between victims and perpetrators : (1) female characters appear more often as victims, and (2) perpetrators are more likely to be White if the victim is Black or Latino. To date, we are the first to show that language used in movie scripts is a strong indicator of violent content, and that there are systematic portrayals of certain demographics as victims and perpetrators in a large dataset. This offers novel computational tools to assist in creating awareness of representations in storytelling.

## 1. INTRODUCTION

Violence is an important narrative tool that can be used to enhance a viewer's experience, boost movie profits and facilitate global market reach [1-3]. Including violent content may modify a viewer's perception of how exciting a movie is by intensifying the sense of relief when a plot-line is resolved favorably [3,4]. However, including violent content in a movie often presents filmmakers with a trade-off between economic advantages and social responsibility. The impact of portrayed violence on the society, especially children and young adults, has been long studied (see detailed review [5]). For example, at an individual level, violent media can increase aggressive affect and aggressive behaviors such as bullying, insulting, pushing and shoving [6,7]. At a societal level, it affects by cultivating the perception of the world as a dangerous place [6]. It may also contribute to the creation of negative stereotypes. For example, by consistently casting people of a particular demographic, it may inflate the perception of that demographic being dangerous [9]; which in time might lead to misrepresentations and societal marginalization [10,11]. This makes studying violence in movies at scale of particular importance.

Most studies that have examined the effect and prevalence of violence in the media have generally used trained human annotators to identify violent content. This approach limits the sample size studied (typically $< 100$), [12,13]. A similar approach has been used to study the relation between a character's demographics and their participation in violent acts. For example, showing that women are more frequently portrayed as victims of violence, while most of the perpetrators are portrayed by middle-aged white actors [13-15]. Automated methods to identify violent content have been limited to audio and video-based classifiers [16,17]. Most similar to our work is that of Martinez et al. [18] which predicts if a movie is violent or not through a deep-learning system based on character's dialogues.

Our objective in this work is to understand the relation between language use in movie scripts and the portrayed violence with respect to a character's demographics. In order to study this relationship, we present experiments to computationally model character violence using features that capture lexical, semantic, sentiment and abusive language characteristics.

## 2. DATA AND METHOD

For the large-scale analysis of violent stereotypes, we obtained Scriptbase [19]. This dataset has been previously studied in the context of movie content analysis [19] as well as for quantifying a character's agency and power [20]. Scriptbase contains 912 Hollywood movie scripts from 23 genres over the years 1909-2013. Each character's utterance has been automatically processed for tagging, parsing, named entity recognition and coreference resolution.

Consistent with Martinez et al. [18], we represent a movie script as a sequence of characters speaking one after another. The violence ratings at the movie-level were obtained from Common Sense Media[1] which have three classes (LOW, MED, HIGH) as described in Martinez et. al.,[18]. We collect language features for each of the character utterances. Our model relies only on pre-trained semantic and sentiment representations for each utterance, plus movie genre. We found that this model uses fewer parameters to achieve better results for the task of classifying movie-level violence ratings (compared to [20]). Semantic representations come from Sent2Vec models [21] pre-trained on a set of 6,000 movie and TV scripts. Sentiment representations are obtained from bidirectional long short-term memory models (LSTM) [22] pre-trained on Stanford Sentiment Treebank [23]. These utterance-representations are fed into a Attention-Based Recurrent Neural Network architecture (RNN) [24] to capture conversational context (i.e., what is being said in relation to what has been previously said). Movie genre binary encoded representation is concatenated to the output of the attention layer . This allows our model to learn that some utterances that are violent for a particular genre may not be considered violent in other genres.

RNN models were implemented in Keras[2]. We used the Adam optimizer with mini-batch size of 16 and learning rate of $0.001$. To prevent overfitting, we use drop-out of 0.5, and train until convergence (i.e., consecutive loss with less than $10^{-8}$ difference). For the RNN layer, we evaluated Gated Recurrent Units [14]. Parameter optimization and estimation of model performance were estimated in a 5-fold cross-validated fashion varying the number of hidden units $H \in [4, 8, 16, 32]$.

From the RNN model we obtain posterior probability of violence for each one of the utterances. For a particular utterance $u_t$, we construct context windows of $k/2$ leading up to and after $u_t$. When required, context windows are padded on either side with empty sentences. This window is then used as input for the RNN model. This procedure yields a sequence of probability estimations over overlapping windows of size $k-1$. $k$ was set to 500 which is on average one-third of the length of a screenplay. This was a proxy to partition a movie into three acts.

---

[1] https://www.commonsensemedia.org/
[2] https://keras.io/

We classified characters as victims, perpetrators or narrators by analyzing the subject-verb-object (SVO) triplets for each utterance. We inspect subjects and objects of actions to identify who (*perpetrator*) is being violent, and towards whom (*victim*). We also distinguish when a character is speaking about violence as a *narrator* (e.g., they attacked her'). These are constructed using the part-of-speech and parsing trees, obtained from spacy[3].

## 3.     RESULTS AND DISCUSSION

In this section we present the results of the large-scale analysis of character-level violence portrayals and the interaction between victims and perpetrators. On all subsequent analysis, we excluded utterances predicted with a LOW violence label, since the concepts of victim and perpetrator are ill-defined for utterances that are not deemed violent.

The violence classification model had a macro F1 score of 0.61 for the three class problem. Semantic form extraction from the utterances classified as MED and HIGH in violence resulted in $1,600,726$ triplets. The four most common forms of SVOs were *pronoun-verb-noun* ($19.03\%$), *pronoun-verb-pronoun* ($8.85\%$), *noun-verb-noun* ($4.60\%$) and p*ronoun-verb-proper noun* ($2.23\%$). From these, we only considered *pronoun-verb-pronoun* and *pronoun-verb-proper* noun since (i) they convey the direction of the action (i.e., from subject to object) and, (ii) both subject and object are most likely to be movie characters. Thus, our analysis focused on these $177,164$ triplets ($11.06\%$).

 In a first analysis, we inspect the number of times an actor participates in violent dialogues based on their demographics. Because a single character can have multiple utterances, we control for this influence by regressing out the speaker ID as a fixed effect in a linear mixed effects model. The dependent variable here is the previously classified victim/perpetrator/narrator. After correction, we test for differences in the frequency of the roles using t-test for gender and ANOVA model for race. Our results show that female characters are more frequently portrayed as victims (HIGH violence $\mathbf{H}_v$: $t(10358) = 2.28, p < 0.05$, MED $\mathbf{M}_v$: $t(74592) = 4.77, p < 0.001$). Additionally, male characters appear more frequently as the perpetrators ($\mathbf{H}_v : t(10723) = -2.67, p < 0.005$, $\mathbf{M}_v : t(77564) = -3.70, p < 0.001$). There was no significant difference in the frequency of narrators ($p \approx 0.3$), suggesting that women and men are equally likely to talk about violence. We did not observe any significant difference in role portrayal based on race (ANOVA, $\mathbf{M}_v : p > 0.05$, $\mathbf{H}_v : p > 0.05$).

In a second analysis, we investigate differences in the way characters interact with each other in the presence of violence. We define a character interaction as a pair of victims and perpetrators. These pairs can be analyzed by gender (e.g., a female character being victimized by another female character) or by race (e.g., a Latino character being violent towards a white character). We compared the frequency of interactions using an omnibus test (ANOVA) and post-hoc pairwise proportion tests with Bonferroni correction. The results of the omnibus test showed that variance in interactions can be explained as a function of the character's race and gender ($F(29, 3062) = 4.658, p < 0.001$). Our results show that female victims interacting with male perpetrators are significantly more frequent--compared to women perpetrating violence either toward men ($\chi^2(1) = 11.67, p < 0.001$) or women ($\chi^2(1) = 14.02, p < 0.001$). The most common interaction was black characters being violent towards characters of mixed race ($z = 0.65$). White characters were shown to be more violent towards Latino ($z = 0.54$) and towards black characters ($z = 0.51$).

---

[3] https://spacy.io/

## 4. CONCLUSION

Our work is the first to study systematic portrayals of certain demographics as victims and perpetrators in a large dataset of movie scripts. Our results on violent content analysis at scale showed that females are more frequently portrayed as subjects of violence, whereas male characters are portrayed as perpetrators of violence. This result is consistent with previous research on media studies [13,15]. Moreover, similar studies on movies and TV have suggested that white male characters are most frequently portrayed as perpetrators [9,13,14]. Complementing these studies, our results suggest that the frequency of White male perpetrator portrayals is conditioned on the victim's race. For example, we found that the agents of violence are more likely to be White if their violence is directed toward Black and Latino characters, but the perpetrators are more likely to be Black if the victim is of Mixed race. This suggests that future studies of stereotypes of violence should consider not only the source of the violence but also the target.